\documentclass[10pt,twocolumn,letterpaper]{article}

\usepackage{iccv}
\usepackage{times}
\usepackage{epsfig}
\usepackage{graphicx}
\usepackage{amsmath}
\usepackage{amssymb}
\usepackage{ragged2e}

\usepackage[breaklinks=true,bookmarks=false]{hyperref}
\usepackage[accsupp]{axessibility}  

\iccvfinalcopy 

\ificcvfinal\fi

\begin{document}

\title{Fantasia3D: Disentangling Geometry and Appearance for High-quality Text-to-3D Content Creation}

\author{Rui Chen\footnotemark[1], Yongwei Chen\footnotemark[1], Ningxin Jiao, Kui Jia\footnotemark[2] \\
South China University of Technology }




\maketitle
\ificcvfinal\thispagestyle{empty}\fi
\def\thefootnote{*}\footnotetext{Equal contribution.}
\def\thefootnote{\dag}\footnotetext{Corresponding author.}
\begin{abstract}
Automatic 3D content creation has achieved rapid progress recently due to the availability of pre-trained, large language models and image diffusion models, forming the emerging topic of text-to-3D content creation. Existing text-to-3D methods commonly use implicit scene representations, which couple the geometry and appearance via volume rendering and are suboptimal in terms of recovering finer geometries and achieving photorealistic rendering; consequently, they are less effective for generating high-quality 3D assets. In this work, we propose a new method of Fantasia3D for high-quality text-to-3D content creation. Key to Fantasia3D is the disentangled modeling and learning of geometry and appearance. For geometry learning, we rely on a hybrid scene representation, and propose to encode surface normal extracted from the representation as the input of the image diffusion model. For appearance modeling, we introduce the spatially varying bidirectional reflectance distribution function (BRDF) into the text-to-3D task, and learn the surface material for photorealistic rendering of the generated surface. Our disentangled framework is more compatible with popular graphics engines, supporting relighting, editing, and physical simulation of the generated 3D assets. We conduct thorough experiments that show the advantages of our method over existing ones under different text-to-3D task settings. Project page and source codes: \url{https://fantasia3d.github.io/}.
\end{abstract}

\begin{figure}
\begin{center}
\includegraphics[width=0.45\textwidth]{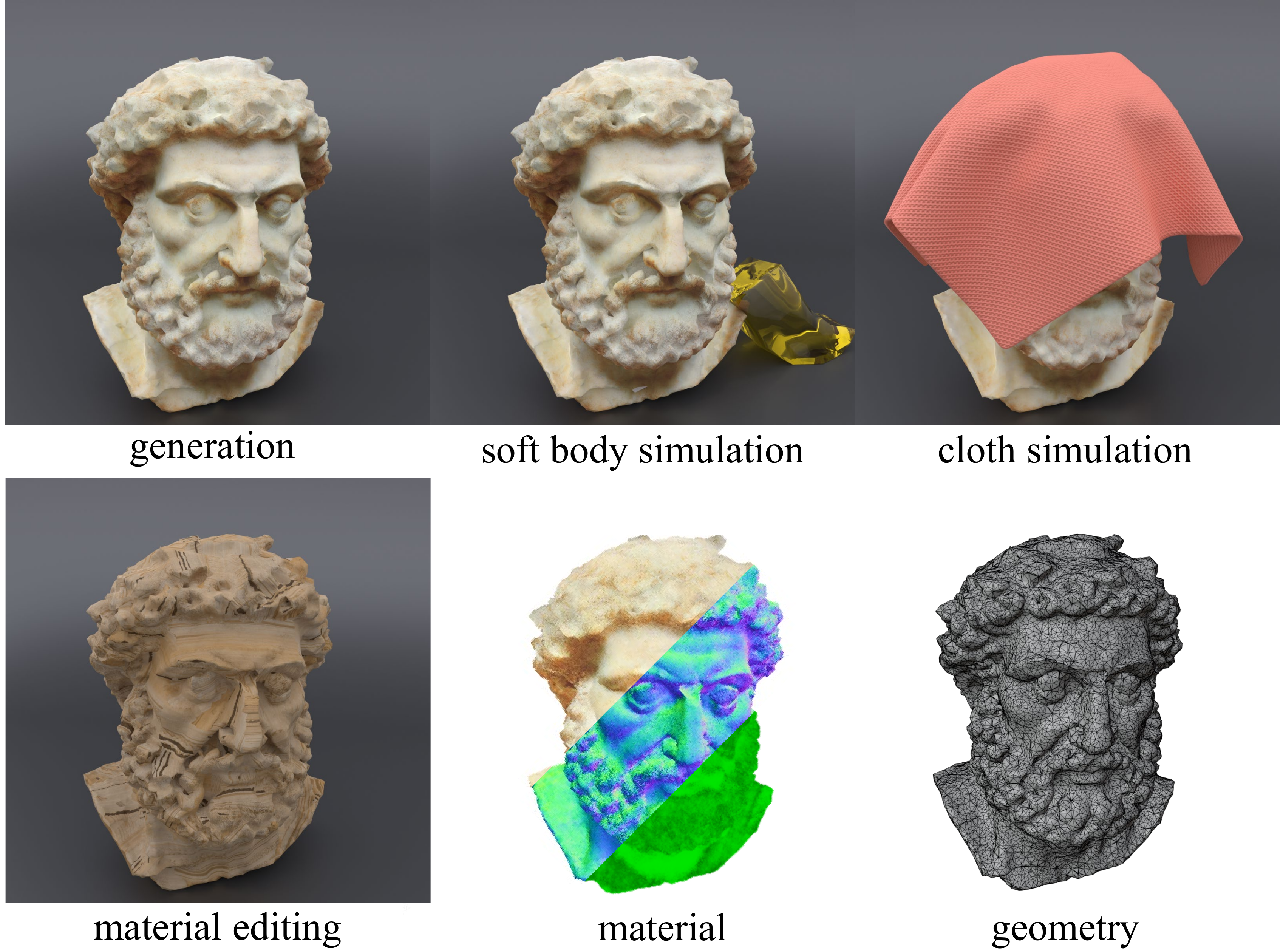}
\end{center}
   \caption{Provided with a textual description of ``a highly detailed stone bust of Theodoros Kolokotronis", our method produces high-quality geometry as well as disentangled materials, and enables photorealistic rendering. }
\label{fig:head}
\end{figure}
\section{Introduction}

Automatic 3D content creation \cite{wang2021neus,lin2022magic3d,poole2022dreamfusion,wang2022rodin} powered by large language models has drawn significant attention recently, due to its convenience to entertaining and gaming industries, virtual/augmented reality, and robotic applications.
The traditional process of creating 3D assets typically involves multiple, labor-intensive stages, including geometry modeling, shape baking, UV mapping, material creation, and texturing, as described in \cite{labschutz2011content}, where different software tools and the expertise of skilled artists are often required. Imperfections would also accumulate across these stages, resulting in low-quality 3D assets. It is thus desirable to automate such a process, and ideally to generate high-quality 3D assets that have geometrically fair surfaces, rich materials and textures, and support photorealistic rendering under arbitrary views.

In this work, we focus on automatic 3D content creation given text prompts encoded by large language models, i.e., the text-to-3D tasks \cite{lin2022magic3d,poole2022dreamfusion}. Text-to-3D is inspired by the tremendous success of text-to-image research \cite{ramesh2022dalle2, saharia2022imagen, nichol2021glide, rombach2022stablediffusion}. To enable 3D generation, most existing methods \cite{poole2022dreamfusion, metzer2022latent-nerf} rely on the implicit scene modeling of Neural Radiance Field (NeRF) \cite{mildenhall2021nerf, Barron2021mipnerf, muller2022instantngp}, and learn the NeRFs by back-propagating the supervision signals from image diffusion models. However, NeRF modeling is less effective for surface recovery \cite{wang2021neus,yariv2021volsdf}, since it couples the learning of surface geometry with that of pixel colors via volume rendering. Consequently, 3D creation based on NeRF modeling is less effective for recovery of both the fine surface and its material and texture. In the meanwhile, explicit and hybrid scene representations \cite{wang2021neus,yariv2021volsdf,shen2021dmtet} are proposed to improve over NeRF by modeling the surface explicitly and performing view synthesis via surface rendering.

In this work, we are motivated to use 3D scene representations that are more amenable to the generation of high-quality 3D assets given text prompts. We present an automatic text-to-3D method called Fantasia3D. Key to Fantasia3D is a disentangled learning of geometry and appearance models, such that both a fine surface and a rich material/texture can be generated. To enable such a disentangled learning, we use the hybrid scene representation of \textsc{DMTet} \cite{shen2021dmtet}, which maintains a deformable tetrahedral grid and a differentiable mesh extraction layer; deformation can thus be learned through the layer to explicitly control the shape generation. For geometry learning, we technically propose to encode a rendered normal map, and use shape encoding of the normal as the input of a pre-trained, image diffusion model; this is in contrast different from existing methods that commonly encode rendered color images. For appearance modeling, we introduce, for the first time, the spatially varying Bidirectional
Reflectance Distribution Function (BRDF) into the text-to-3D task, thus enabling material learning that supports photorealistic rendering of the learned surface. We implement the geometry model and the BRDF appearance model as simple MLPs. Both models are learned through the pre-trained image diffusion model, using a loss of Score Distillation Sampling (SDS) \cite{poole2022dreamfusion}. We use the pre-trained stable diffusion \cite{rombach2022stablediffusion,stablediffusion_code} as the image generation model in this work.

We note that except for text prompts, our method can also be triggered with additional inputs of users' preferences, such as a customized 3D shape or a generic 3D shape of a certain object category; this is flexible for users to better control what content is to be generated. In addition, given the disentangled generation of geometry and appearance, it is convenient for our method to support relighting, editing, and physical simulation of the generated 3D assets. We conduct thorough experiments to verify the efficacy of our proposed methods. Results show that our proposed Fantasia3D outperforms existing methods for high-quality and diverse 3D content creation. We summarize our technical contributions as follows.

\begin{itemize}
\item We propose a novel method, termed Fantasia3D, for high-quality text-to-3D content creation. Our method disentangles the modeling and learning of geometry and appearance, and thus enables both a fine recovery of geometry and photorealistc rendering of per-view images. 
\item For geometry learning, we use a hybrid representation of \textsc{DMTet}, which supports learning surface deformation via a differentiable mesh extraction; we propose to render and encode the surface normal extracted from \textsc{DMTet} as the input of the pre-trained image diffusion model, which enables more subtle control of shape generation.
\item For appearance modeling, to the best of our knowledge, we are the first to introduce the full BRDF learning into text-to-3D content creation, facilitated by our proposed geometry-appearance disentangled framework. BRDF modeling promises high-quality 3D generation via photorealistic rendering.
\end{itemize}

\begin{figure*}
\begin{center}
\includegraphics[width=0.9\textwidth]{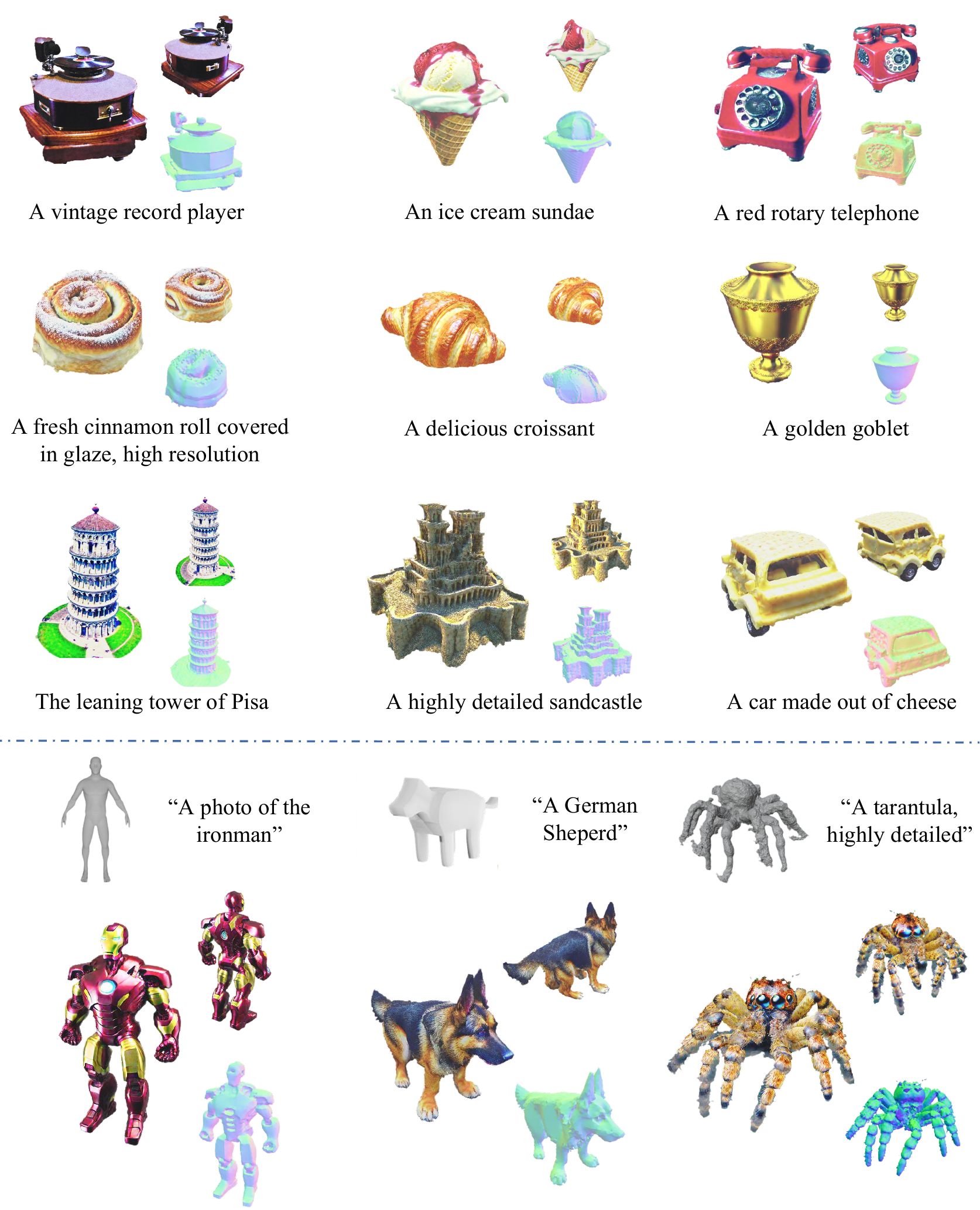}
\end{center}
\vspace{-0.5em}
   \caption{\textbf{Results of our method.} The upper portion of this figure showcases
 the generation results obtained from solely text prompts. The lower portion showcases user-guided generation results given guiding meshes with the corresponding textual descriptions.}
\label{fig:main_results}
\end{figure*}

\begin{figure*}
\begin{center}
\includegraphics[width=0.98\textwidth]{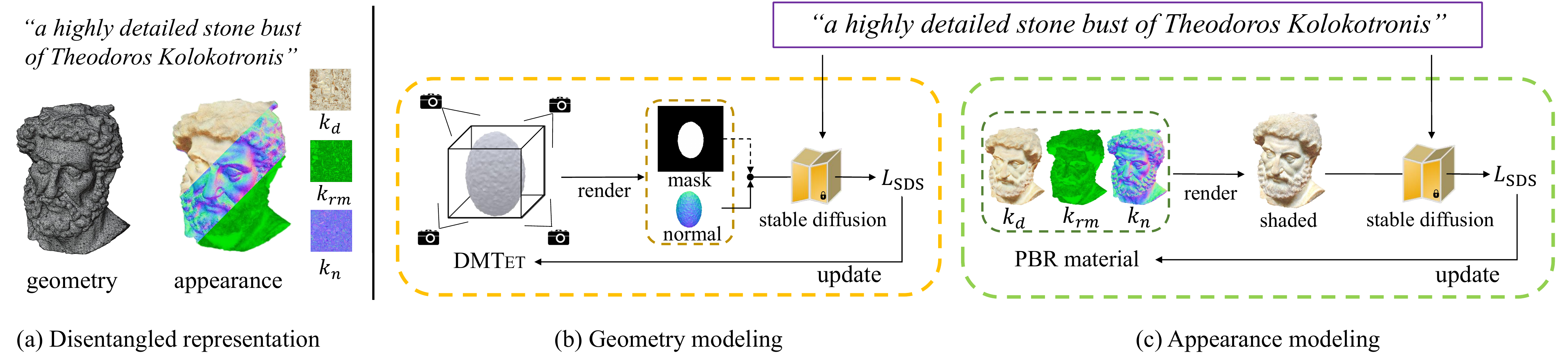}
\end{center}
   \caption{\textbf{Overview of our method.} Our method can generate disentangled geometry and appearance given a text prompt (cf. figure (a)), which are produced by (b) geometry modeling and (c) appearance modeling, respectively. (b) We employ \textsc{DMTet} as our 3D geometry representation, which is initialized as a 3D ellipsoid here. To optimize the parameters of \textsc{DMTet}, we render the normal map (and the object mask in the early training phase) of the extracted mesh from DMTet as the shape encoding of stable diffusion \cite{rombach2022stablediffusion,stablediffusion_code}. (c) For appearance modeling, we introduce the spatially-varying Bidirectional Reflectance Distribution Function (BRDF) modeling into text-to-3D generation, and learn to predict three components (namely, $k_d$, $k_{rm}$, and $k_n$) of the appearance. Both geometry and appearance modeling are supervised by Score Distillation Sampling (SDS) loss \cite{poole2022dreamfusion}. }
\label{fig:pipeline}
\end{figure*}

\section{Related work}
 \noindent
\textbf{Text-to-3D content creation.} Motivated by the desire to generate high-quality 3D content from simple semantics such as text prompts, text-to-3D has drawn considerable attention in recent years \cite{poole2022dreamfusion,jain2022dreamfiled,lin2022magic3d}. Existing methods either use pre-trained 2D text-to-image models \cite{rombach2022stablediffusion,balaji2022eDiff-I,saharia2022imagen}, together with score distillation sampling \cite{poole2022dreamfusion}, to generate 3D geometries \cite{lin2022magic3d} or synthesize novel views \cite{poole2022dreamfusion}, or train a text-conditioned 3D generative model from scratch \cite{shue2022triplanediffusion,zhou2021pointvoxeldiffusion,li2022diffusionsdf,nam20223dldm}. These methods generate 3D geometries with little exploration of generating high-quality lighting and surface materials. On the other hand,  TANGO \cite{chen2022tango} is able to generate high-quality surface materials given text prompts; unfortunately, the method requires as input a 3D surface mesh. Our proposed method addresses the shortcomings of the above methods, and is able to generate both high-quality surface geometries and their corresponding materials, both of which are crucial for photorealistic rendering of the generated 3D content. Our method thus, for the first time, closes the loop of object-level text-to-3D content creation. 

\noindent\textbf{Surface material estimation.} The estimation of surface materials is a long-standing challenge in computer vision and graphics research. Earlier methods \cite{aittala2013practical,wu2011sparse} focus on recovering physically based materials under known lighting conditions, whose usefulness is, however, limited in real-world scenarios. Subsequent methods \cite{gao2019deepbrdf,deschaintre2018single,bi2020neural,aittala2016reflectance} try to estimate materials under natural lighting conditions, assuming the availability of complete geometry information. More recently, the joint reconstruction of geometry and materials is proposed given calibrated multi-view images \cite{chen2019learning,chen2021dib,liu2019soft,zhang2020image}. Alternative to these methods, we explore the novel creation of surface materials and geometries from trained language models.

\section{Preliminary}
In this section, we present a few preliminaries that are necessary for presenting our proposed method in Section \ref{sec:Method}.

\subsection{Score distillation sampling}
\label{sec:sds}
DreamFusion \cite{poole2022dreamfusion} presents a method that optimizes 3D scene parameters and synthesizes novel views from textual descriptions, by employing a pre-trained 2D diffusion model. The scene is represented as a differentiable image parameterization \cite{mordvintsev2018dip}, where a differentiable generator $g$ renders 2D images $x = g(\theta)$ from a modified Mip-NeRF \cite{Barron2021mipnerf} parameterized as $\theta$. DreamFusion leverages a diffusion model $\phi$ (Imagen \cite{saharia2022imagen} in this instance) to provide a score function $\hat{\epsilon}_{\phi}(x_t;y,t)$, which predicts the sampled noise $\epsilon$ given the noisy image $x_t$, text-embedding $y$, and noise level $t$. This score function guides the direction of the gradient for updating the scene parameters $\theta$, and the gradient is calculated by Score Distillation Sampling (SDS): 
\begin{equation}\label{EqnSDS}
    \bigtriangledown_\theta \mathcal{L}_{\text{SDS}}(\phi,x) = \mathbb{E}_{t,\epsilon} \left [ w(t)(\hat{\epsilon}_{\phi}(x_t;y,t) - \epsilon)\frac{\partial x}{\partial \theta} \right ],
\end{equation}
while $w(t)$ is a weighting function. Since Imagen is not publicly accessible, in this work, we use the released latent space diffusion model of Stable Diffusion \cite{rombach2022stablediffusion,stablediffusion_code} as our guidance model, and revise the SDS loss (\ref{EqnSDS}) accordingly. Details are given in Section \ref{sec:Geometry Optimization}.

\subsection{\textbf{\textsc{DMTet}}}
\label{DMTet}
Implicit surface representations \cite{mildenhall2021nerf,park2019deepsdf,mescheder2019occupancy} are popularly used in novel view synthesis and 3D reconstruction, due to their capabilities to represent complex scenes. However, surfaces of lower quality may be obtained \cite{munkberg2022extracting} by extracting explicit meshes from these implicit representations using marching cubes \cite{lorensen1987marchingcube}. Instead, Shen et al. \cite{shen2021dmtet}  propose a hybrid representation, termed \textsc{DMTet}, that has two key features, i.e., a deformable tetrahedral grid and a differentiable Marching Tetrahedral (MT) layer. The deformable tetrahedral grid $(V_T,T)$ has vertices $V_T$ in the tetrahedral grid $T$. For each vertex $v_i \in V_T$, the proposed method predicts the Signed Distance Function (SDF) value $s(v_i)$ and a position offset $\triangle v_i$ by:
\begin{equation}
    (s(v_i),\triangle v_i) = \Psi(v_i;\psi),
    \label{eq:dmtet}
\end{equation}
where $\psi$ is the parameters of a network $\Psi$, enabling the extraction of explicit meshes through MT layer during each iteration of training. In this work, We use \textsc{DMTet} as our geometry representation and render the mesh extracted from MT layer iteratively by a differentiable renderer\cite{Laine2020diffrast,munkberg2022extracting}.

\section{The Proposed Method}
\label{sec:Method}
In this section, we present our proposed method of Fantasia3D for high-quality text-to-3D object generation, by disentangling the modeling and learning of geometry and appearance.
For geometry modeling, we rely on the hybrid surface representation of \textsc{DMTet}, and parameterize the 3D geometry as an MLP $\Psi$ that learns to predict the SDF value and position offset for each vertex in the deformable tetrahedral grid of \textsc{DMTet}; in contrast to previous methods, we propose to use the rendered normal map (and the object mask in the early training phase) of the extracted mesh from \textsc{DMTet} as the input of shape encoding. For appearance modeling, we introduce, for the first time, the full Bidirectional
Reflectance Distribution Function (BRDF) modeling into text-to-3D generation, and learn an MLP $\Gamma$ that predicts parameters of surface material and supports high-quality 3D generation via photorealistic rendering.
Given the disentangled modeling of $\Psi$ and $\Gamma$, the whole pipeline is learned with SDS supervision, and the gradients are back-propagated through the pre-trained stable diffusion model.
Our pipeline is initialized either as a 3D ellipsoid or as a customized 3D model provided by users. Fig. \ref{fig:pipeline} gives an illustration of our proposed method.
In contrast, previous methods couple the geometry and appearance learning, and are suboptimal in terms of leveraging the powerful pre-trained 2D image diffusion models via SDS loss.
Details of the proposed Fantasia3D are presented as follows.
 
\subsection{\textbf{\textsc{DMTet}} initialization}
We adopt \textsc{DMTet} as our 3D scene representation, which is parameterized as the MLP network $\Psi$. For each vertex $v_i \in V_{T}$  of the tetrahedral grid $(V_{T},T)$, $\Psi$ is trained to predict the SDF value $s(v_i)$ and the deformation offset $\triangle v_i$. A triangular mesh can be extracted from $(V_{T},T)$ using the MT layer, whose procedure is also differentiable w.r.t. the parameters of $\Psi$.
We initialize \textsc{DMTet} either as a 3D ellipsoid or as a customized 3D model provided by users; the latter choice is useful when the text-to-3D task is to be conditioned on users' preferences. In either case, we initialize $\Psi$ by the following fitting procedure. We sample a point set $\{ p_i \in \mathbb{R}^3 \}$ whose points are in close proximity to the initialized 3D ellipsoid or customized model, and compute their SDF values, resulting in $\{ SDF (p_i) \}$; we use the following loss to optimize the parameters $\psi$ of $\Psi$:
\begin{equation}
     \mathcal{L}_\text{SDF}=\sum_{p_i\in P} \left \| s(p_i;\psi) -SDF(p_i)\right \|_{2}^{2} .
\end{equation}

\subsection{Geometry modeling}
\label{sec:Geometry Optimization}

Previous text-to-3D methods \cite{poole2022dreamfusion,jain2022dreamfiled} commonly use NeRF \cite{mildenhall2021nerf,Barron2021mipnerf} as the implicit scene representation, which couples the geometry with color/appearance and use volume rendering for view synthesis. Since NeRF modeling is less effective for surface reconstruction \cite{wang2021neus,yariv2021volsdf}, these methods are consequently less effective to generate high-quality 3D surfaces by back-propagating the supervision of SDS loss through pre-trained text-to-image models. As a remedy, the method \cite{lin2022magic3d} uses a second stage of the refined generation that is based on scene modeling of explicit surfaces.

In this work, we propose to decouple the generation of geometry from that of appearance, based on the hybrid scene representation of \textsc{DMTet}, which enables photorealistic surface rendering to make better use of the powerful pre-trained text-to-image models. More specifically, given the current \textsc{DMTet} with MLP parameters $\psi$, we generate a normal map $n$, together with an object mask $o$, as:
\begin{equation}
    (n,o) = g_n(\psi,c),
    \label{eq:image}
\end{equation}
where $g_n$ is a differentiable render (\eg, nvidiffrast \cite{Laine2020diffrast}), and $c$ is a sampled camera pose. We randomly sample the camera poses in the spherical coordinate system to ensure that the camera poses are distributed uniformly on the sphere.
We propose to use the generated $n$ (and $o$) as the input of shape encoding to connect with stable diffusion. To update $\psi$, we again employ SDS loss that computes the gradient w.r.t. $\psi$ as:
\begin{equation}\label{sds_geometry}
    \bigtriangledown_\psi \mathcal{L}_{\text{SDS}}(\phi, \tilde{n}) = \mathbb{E} \left [ w(t)(\hat{\epsilon}_{\phi}(z_t^{\tilde{n}};y,t) - \epsilon)\frac{\partial \tilde{n}}{\partial \psi} \frac{{\partial z^{\tilde{n}}}}{\partial \tilde{n}}\right ],
\end{equation}
where $\phi$ parameterizes the pre-trained stable diffusion model, $\tilde{n}$ denotes concatenation of the normal $n$ with the mask $o$, $z^{\tilde{n}}$ is the latent code of $\tilde{n}$ via shape encoding, $\hat{\epsilon}_{\phi}(z_t^{\tilde{n}};y,t)$ is the predicted noise given text embedding $y$ and noise level $t$, and $\epsilon$ is the noise added in $z_t^{\tilde{n}}$.
In practice, we utilize a coarse-to-fine strategy to model the geometry. During the early phase of training, we use the downsampled $\tilde{n}$ as the latent code, which is inspired by \cite{metzer2022latent-nerf}, to rapidly update $\Psi$ and attain a coarse shape. However, a domain gap exists between $\tilde{n}$ and the latent space data distribution learned by the VAE encoder in the stable diffusion, which may lead to a mismatch of the generated geometry from the textual description. To mitigate this discrepancy, we implement a data augmentation technique by introducing random rotations to $\tilde{n}$. Our experimental observations reveal that this technique enhances the alignment between the generated geometry and the provided textual description. In the later phase of training, aiming to capture finer geometric details with greater precision, we encode the high-resolution normal $n$ (without the mask $o$) to derive $z^n$, using the pre-trained image encoder in stable diffusion.

\begin{figure}
\begin{center}
\includegraphics[width=0.45\textwidth]{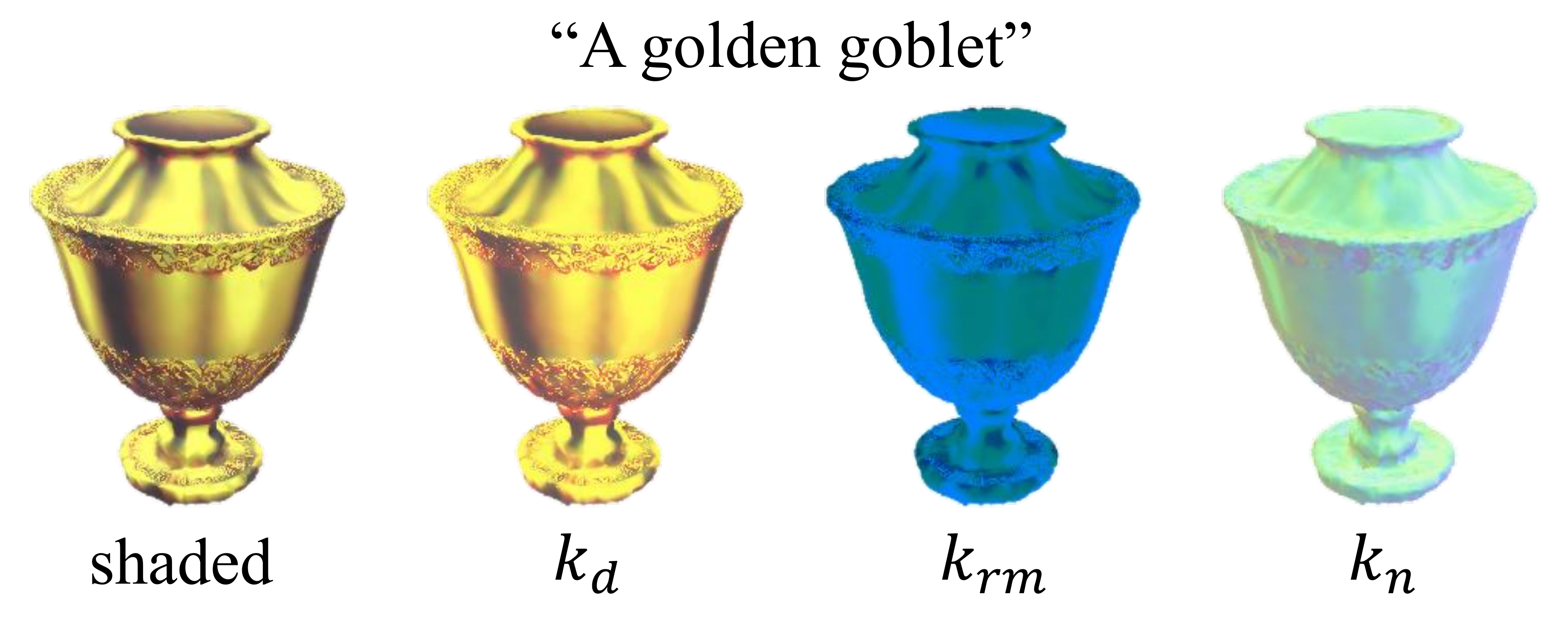}
\vspace{-1em}
\end{center}
   \caption{Three components of the material model, namely the diffuse term $k_d$, the roughness and metallic term $k_{rm}$, and the normal variation term $k_{n}$. }
\label{fig:brdf decomposition}
\end{figure}

\begin{figure}
\begin{center}
\includegraphics[width=0.45\textwidth]{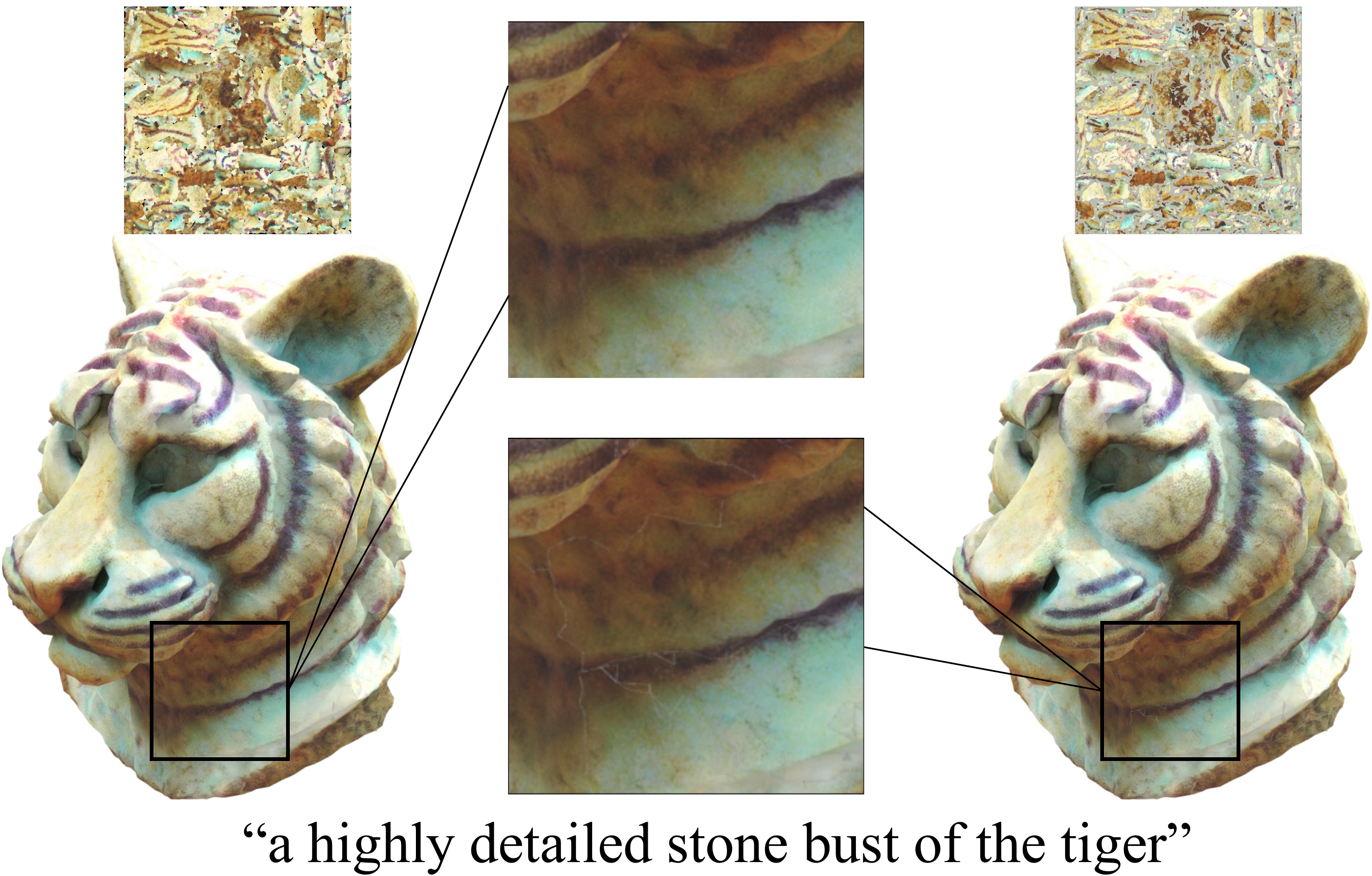}
\end{center}
   \caption{A comparison of UV edge padding (cf. the left column) and original texturing (cf. the right column). UV edge padding removes the white seams that appear in the right rendering.}
\label{fig:seam}
\end{figure}

\subsection{Appearance modeling}
\label{sec:BRDF Decomposition}

Given a learned \textsc{DMTet} with geometry parameters $\psi$, we aim for photorealistic surface rendering to better leverage the powerful image diffusion model. This is achieved by introducing into text-to-3D generation the Physically-Based Rendering (PBR) material model. As illustrated in Fig. \ref{fig:brdf decomposition}, the material model we use \cite{mcauley2012practical} comprises three components, namely the diffuse term $k_d\in \mathbb{R}^3$, the roughness and metallic term $k_{rm} \in \mathbb{R}^2$, and the normal variation term $k_n \in \mathbb{R}^3$. The $k_{d}$ term denotes the diffuse value, while the $k_{rm}$ term encompasses the roughness $r$ and metalness $m$; the roughness $r$ serves as a metric for measuring the extent of specular reflection and a parameter of the GGX \cite{walter2007microfacet} normal distribution function in the rendering equation. The metalness parameter $m$ and diffuse value $k_d$ can be used to compute the specular term $k_s$ using $k_s=(1-m) \cdot 0.04+m \cdot k_{d}$. Furthermore, we use a normal variation $k_n$ in tangent space to enhance the surface lighting effect and introduce additional geometric variations.

We use an MLP $\Gamma$ as our parameterized material model. $\Gamma$ is learned to predict spatially varying material parameters, which are subsequently used to render the surface extracted from \textsc{DMTet}. More specifically, for any point $p \in \mathbb{R}^{3}$ on the surface, we use hash-grid positional encoding \cite{muller2022instantngp}, and generate the diffuse term $k_{d}$, the specular term $k_{rm}$, and the tangent space normal $k_{n}$ as:
\begin{equation} \label{eq:texture_mapping}
(k_d,k_{rm},k_n) = \Gamma(\beta(p);\gamma),
\end{equation}
where $\beta$ is the positional encoding of $p$, and $\gamma$ parameterizes the MLP $\Gamma$. 
The basic rendering equation suggests that each image pixel at a specific viewing direction can be rendered as  
\begin{equation} \label{eq:rendering_equation}
L(p,\omega)=\int_{\Omega}L_{i}(p,\omega_{i})f(p,\omega_{i},\omega)(\omega_{i} \cdot n_p) \mathrm{d}\omega_{i},
\end{equation}
where $L$ is the rendered pixel color along the direction $\omega$ from the surface point $p$, $\Omega = \left \{ \mathbf{\omega}_i: \mathbf{\omega}_i \cdot n_p \ge 0 \right \}$ denotes a hemisphere with the incident direction $\omega_i$ and surface normal $n_p$ at $p$, $L_i$ is the incident light that is represented by an off-the-shelf environment map \cite{poliigon}, and $f$ is the BRDF determined by the material parameters $(k_d,k_{rm},k_n)$ predicted by (\ref{eq:texture_mapping}).
We note that $L$ is the summation of diffuse intensity $L_{d}$ and specular intensity $L_{s}$, and the two terms can be computed as follows:
\begin{equation} \label{eq:split_sum_all}
\begin{split}
 L(p,\omega)= L_{d}(p) + L_{s}(p,\omega), \\
 L_{d}(p)=k_d(1-m)\int_{\Omega}L_{i}(p,\omega_{i})(\omega_{i} \cdot n_p) \mathrm{d}\omega_{i}, \\
 L_{s}(p,\omega)=\int_{\Omega}\frac{DFG}{4(\omega \cdot n_p)(\omega_{i} \cdot n_p)}L_{i}(p,\omega_{i})(\omega_{i} \cdot n_p)\mathrm{d}\omega_{i},
 \end{split}
 \end{equation}
 where $F$, $G$, and $D$ represent the Fresnel term, the shadowing-mask term, and the GGX distribution of normal, respectively. Following \cite{munkberg2022extracting}, the hemisphere integration can be calculated using the split-sum method.
 
By aggregating the rendered pixel colors along the direction $\omega$ (i.e., camera pose), we have the rendered image $x = \{ L(p,\omega) \}$ that connects with the image encoder of the pre-trained stable diffusion model. We update the parameters $\gamma$ by computing the gradient of the SDS loss w.r.t. $\gamma$:
\begin{equation}\label{EqnGradSDSLossWRTgamma}
    \bigtriangledown_\gamma \mathcal{L}_{\text{SDS}}(\phi,x) = \mathbb{E} \left [ w(t)(\hat{\epsilon}_{\phi}(z_t^x;y,t) - \epsilon)\frac{\partial x}{\partial \gamma} \frac{{\partial z^x}}{\partial x}\right ]. 
\end{equation}
Notations in (\ref{EqnGradSDSLossWRTgamma}) are similarly defined as those in (\ref{sds_geometry}).

\noindent\textbf{Texturing}. Given the trained $\Gamma$, we proceed by sampling the generated appearance as 2D texture maps, in accordance with the UV map generated by the xatlas \cite{xatlas2021}. Note that texture seams would emerge by direct incorporation of the sampled 2D textures into graphics engines (e.g., Blender \cite{blender}). We instead employ the UV edge padding technique \cite{padding}, which involves expanding the boundaries of UV islands and filling empty regions. As illustrated in Fig. \ref{fig:seam}, this padding technique removes background pixels in the texture map and also removes the seams in the resulting renderings.

\noindent\textbf{Implementation Details.} We implement the network $\Psi$ as a three-layer MLP with 32 hidden units, and implement $\Gamma$ as a two-layer MLP with 32 hidden units. Our method is optimized on 8 Nvidia RTX 3090 GPUs for about 15 minutes for learning $\Psi$ and about 16 minutes for learning $\Gamma$, respectively, where we use AdamW optimizer with the respective learning rates of $ 1\times 10^{-3} $ and $ 1\times 10^{-2} $.  For each iteration, we uniformly sample 24 camera poses for the rendering of normal maps and colored images. More details of our implementation are available in the supplemental material. In geometry modeling, we set $\omega(t) =\sigma ^{2}$ during the early phase and then transition to $w(t) = \sigma^2 \sqrt{1-\sigma^2}$ as we progress to the later phase. In appearance modeling, we apply $w(t) = \sigma^2 \sqrt{1-\sigma^2}$ during the early phase, followed by a shift to $1/\sigma^2$ as we enter the later phase. This approach mitigates the issue related to over-saturated color within the appearance modeling.

\section{Experiments}

In this section, we present comprehensive experiments to evaluate the efficacy of our proposed method for text-to-3D content creation. We first conduct ablation studies in Section \ref{exp:ablation_study} that verify the importance of our key design of disentangling geometry and appearance for text-to-3D generation. In Section \ref{exp:zeroshot}, we show the efficacy of our method for the generation of 3D models with PBR materials from arbitrary text prompts, where we also compare with two recent state-of-the-art methods (namely, Magic3D \cite{lin2022magic3d} and DreamFusion \cite{poole2022dreamfusion}). In Section \ref{exp:user-guided}, we present our results under the setting of user-guided generation and compare them with Latent-NeRF \cite{metzer2022latent-nerf}. We finally demonstrate in Section \ref{exp:scene_editing} that the 3D assets generated by our method are readily compatible with popular graphics engines such as Blender \cite{blender}, thus facilitating relighting, editing, and physical simulation of the resulting 3D models.
\begin{figure*}
\begin{center}
\includegraphics[width=0.9\textwidth]{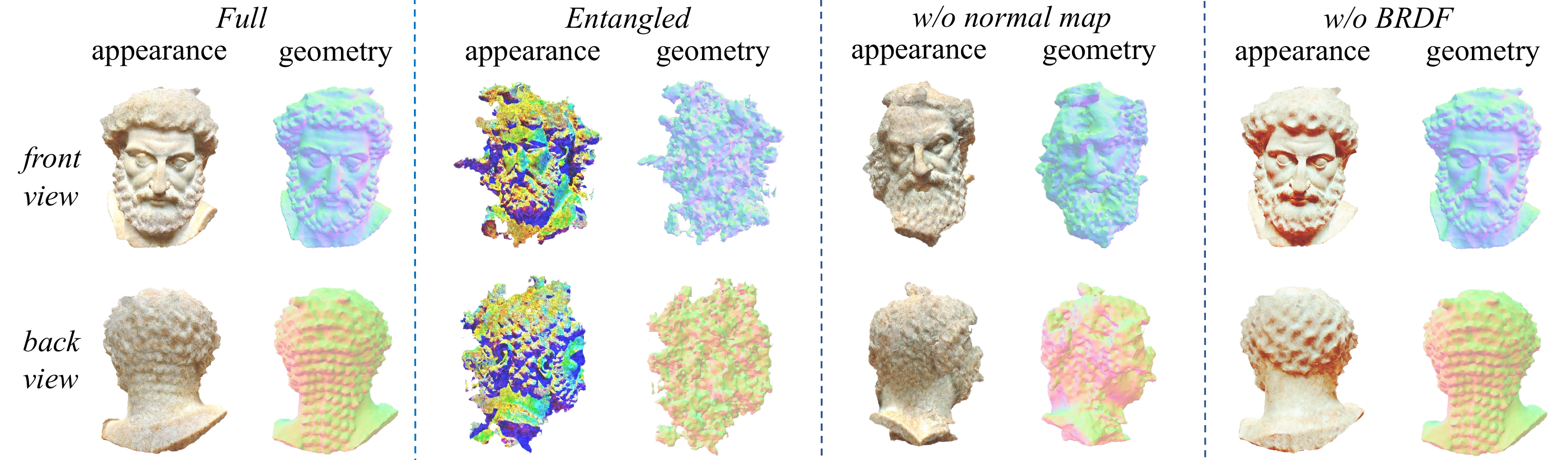}
\end{center}
   \caption{\textbf{Ablation studies of our method.} The text prompt is ''a highly detailed stone bust of Theodoros Kolokotronis". Please refer to Section \ref{exp:ablation_study} for specific settings of individual columns. Please refer to the video results in the supplemental materials for better comparisons.}
\label{fig:ablation}
\end{figure*}

 \begin{figure*}
\begin{center}
\includegraphics[width=0.9\textwidth]{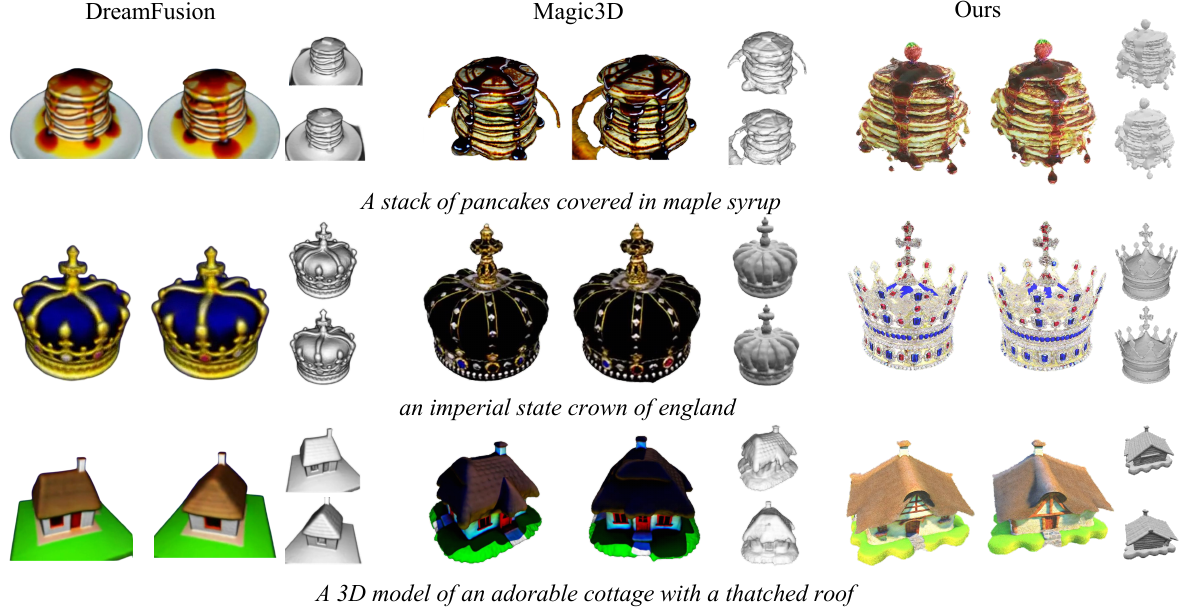}
\end{center}
\vspace{-0.43em}
   \caption{\textbf{Comparison of zero-shot generation.} Since DreamFusion and Magic3D do not have released codes, their results are obtained by downloading from their project pages. More results and comparisons are available in the supplemental materials.}
\label{fig:zero-shot generation}
\end{figure*}

\subsection{Ablation studies}\label{exp:ablation_study}

We use an example text prompt of ``a highly detailed stone bust of Theodoros Kolokotronis'' for the ablation studies. Results of alternative settings are given in Fig. \ref{fig:ablation}. As the reference, the first two columns show the geometry and appearance results of our Fantasia3D rendered from the front and back views.
To verify the effectiveness of the disentangled design in Fantasia3D, we also conduct experiments as follows: in each iteration, we render shaded images of the mesh extracted from \textsc{DMTet}, and learn to update the parameters of a network responsible for both the geometry and material, using the gradient computed by SDS loss. Results in the second two columns show that such an entangled learning fails to generate plausible results.
Previous methods (e.g., DreamFusion \cite{poole2022dreamfusion} and Magic3D \cite{lin2022magic3d}) couple the geometry and appearance generation together, following NeRF \cite{mildenhall2021nerf}. Our adoption of the disentangled representation is mainly motivated by the difference of problem nature for generating surface geometry and appearance. In fact, when dealing with finer recovery of surface geometry from multi-view images, methods (e.g.,  VolSDF \cite{yariv2021volsdf}, nvdiffrec \cite{munkberg2022extracting}, etc) that explicitly take the surface modeling into account triumph; our disentangled representation enjoys the benefit similar to these methods. The disentangled representation also enables us to include the BRDF material representation in the appearance modeling, achieving better photo-realistic 
rendering by the BRDF physical prior.

To investigate how shape encoding of the normal map plays role in Fantasia3D, we replace the normal map with an image that is shaded on the mesh extracted from \textsc{DMTet} using fixed material parameters; results in the third two columns become weird with twisted geometries. This is indeed one of the key factors that makes the success of Fantasia3D. Our initial hypothesis is that shape information contained in normal and mask images could be beneficial to geometry learning, and as such, we further observe that the value range of normal maps is normalized in (-1, 1), which aligns with the data range required for latent space diffusion; our empirical studies verify our hypothesis. Our hypothesis is further corroborated by observing that the LAION-5B \cite{schuhmann2022laion} dataset used for training Stable Diffusion contains normals
(referring to \href{https://rom1504.github.io/clip-retrieval/?back=https%3A%2F%2Fknn.laion.ai&index=laion5B-H-14&useMclip=false&query=normal+map}{website} 
for retrieval of normal data in LAION-5B \cite{schuhmann2022laion}), which allows Stable Diffusion to handle the optimization of normal maps effectively. To deal with rough and coarse geometry in the early stage of learning, we use the concatenated 64 $\times$ 64 $\times$ 4 (normal, mask) images for better convergence. As the learning progresses, it becomes essential to render the 512 $\times$ 512 $\times$ 3 high-resolution normal images for capturing finer geometry details, and we choose to use normal images only in the later stage. This strategy strikes an accuracy-efficiency balance throughout the geometry optimization process.

Finally, we replace the full BRDF in Fantasia3D with a simple diffuse color rendering; the results in the last two columns become less realistic and are short of reflection effects when rendered from different views.

\subsection{Zero-shot generation}
\label{exp:zeroshot}

\begin{figure}[h]
\begin{center}
\includegraphics[width=0.48\textwidth]{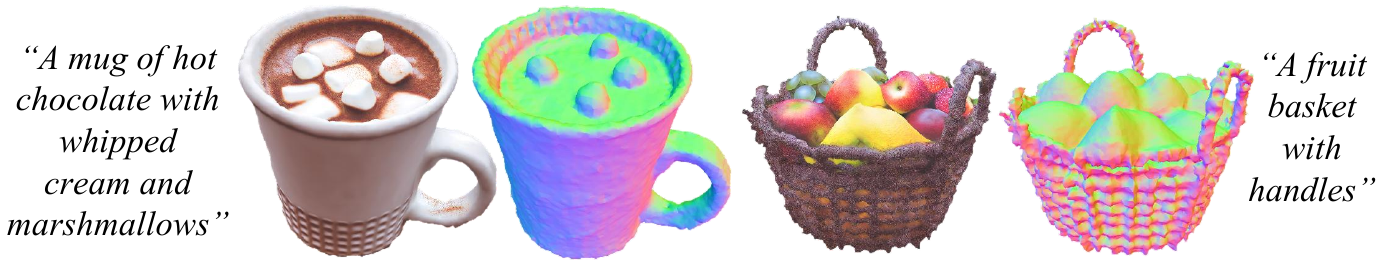}
\end{center}
\caption{\textbf{Geometries beyond genus-zero ones.} \textsc{DMTet} can deform to any topologies, which enables Fantasia3D to generate complex geometries, including those beyond genus-zero ones.}
\label{fig:geometry}
\end{figure}

\begin{figure}[h]
\begin{center}
\includegraphics[width=0.48\textwidth]{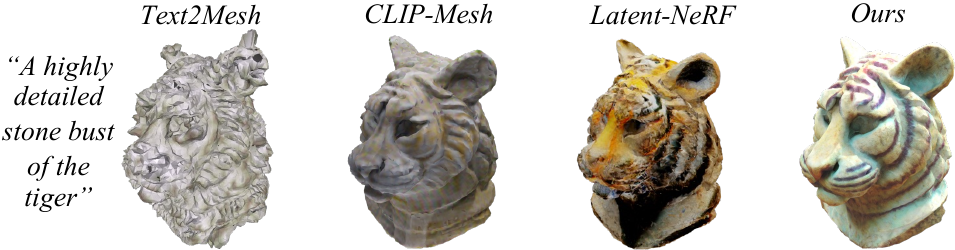}
\end{center}
\caption{\textbf{Comparison of Texturing.}  We compare the appearance stage of Fantasia3D with three text-driven texturing methods, using the geometry generated by the geometry stage of
Fantasia3D. }
\label{fig:textures}
\end{figure}

In this section, we evaluate our method for generating 3D assets from solely natural language descriptions (i.e., the setting of zero-shot generation), by comparing with two state-of-the-art methods, namely DreamFusion \cite{poole2022dreamfusion} and Magic3D \cite{lin2022magic3d}.
Fig. \ref{fig:zero-shot generation} gives the comparative results given the same text prompts. Since DreamFusion and Magic3D do not have released codes, their results are obtained by downloading from their project pages. Comparing our method with Magic3D, we observe that our results are more photorealistic with competitive geometries. We consistently outperform DreamFusion in both appearance and geometry generation. Notably, our method also offers the convenience of easy geometry extraction and editing, as demonstrated in \ref{exp:scene_editing}, which are less obvious from DreamFusion or Magic3D.  More of our results are given in the top half of Fig. \ref{fig:main_results} and Fig. \ref{fig:geometry}. Furthermore, we compare our appearance modeling stage with several mesh stylization methods, namely Text2mesh \cite{text2mesh}, CLIP-Mesh \cite{khalid2022clipmesh} and Latent-NeRF \cite{metzer2022latent-nerf}. Fig. \ref{fig:textures} shows that our method excels in generating more
realistic appearances, outperforming the other competitors. We present additional results and comparisons in the supplemental materials.

\begin{figure}
\begin{center}
\includegraphics[width=0.48\textwidth]{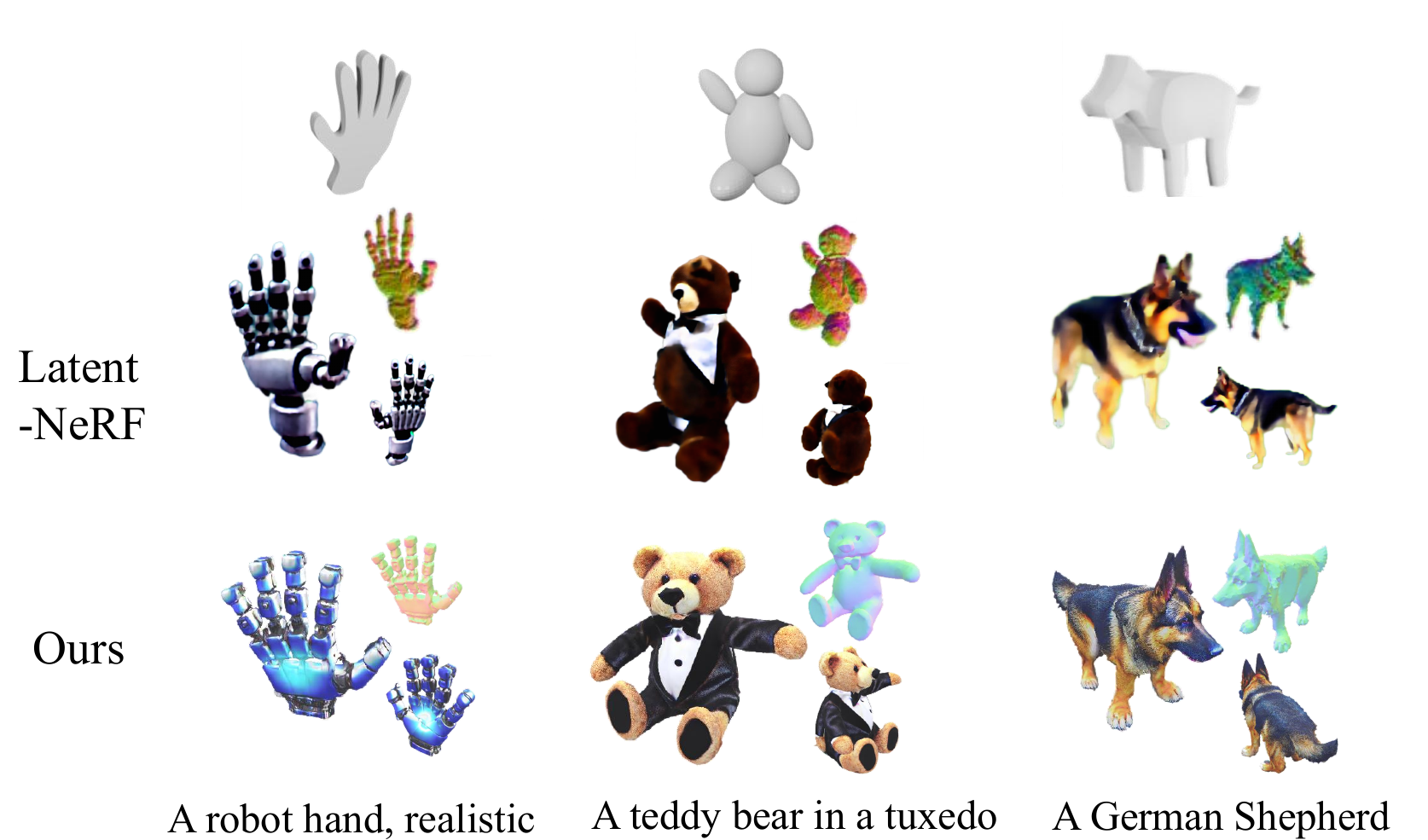}
\end{center}
       \caption{\textbf{Comparison of user-guided generation.} The top row shows the input meshes provided by users, the bottom row gives the input text prompts, and the middle two rows show the results from the comparative methods.}
\label{fig:user-guided generation}
\end{figure}
\subsection{User-guided generation}
\label{exp:user-guided}

In addition to zero-shot generation, our method is flexible to accept a customized 3D model as the initialization, alternative to a 3D ellipsoid, thereby facilitating user-guided asset generation. As shown in the lower half of Fig. \ref{fig:main_results}, our method is able to generate rich details in both the geometry and appearance when provided with low-quality 3D models for initialization. The three meshes used in the experiments are from Text2Mesh \cite{text2mesh}, Latent-NeRF \cite{metzer2022latent-nerf}, and the creation of Stable DreamFusion \cite{stabledreamfusion}, respectively. We also compare with the state-of-the-art approach of Latent-NeRF \cite{metzer2022latent-nerf} under this user-guided generation setting; results are given in Fig. \ref{fig:user-guided generation}. Our method outperforms Latent-NeRF \cite{metzer2022latent-nerf} in both the geometry and texture generation when given the same input meshes and text prompts.

\begin{figure}
\begin{center}
\includegraphics[width=0.41\textwidth]{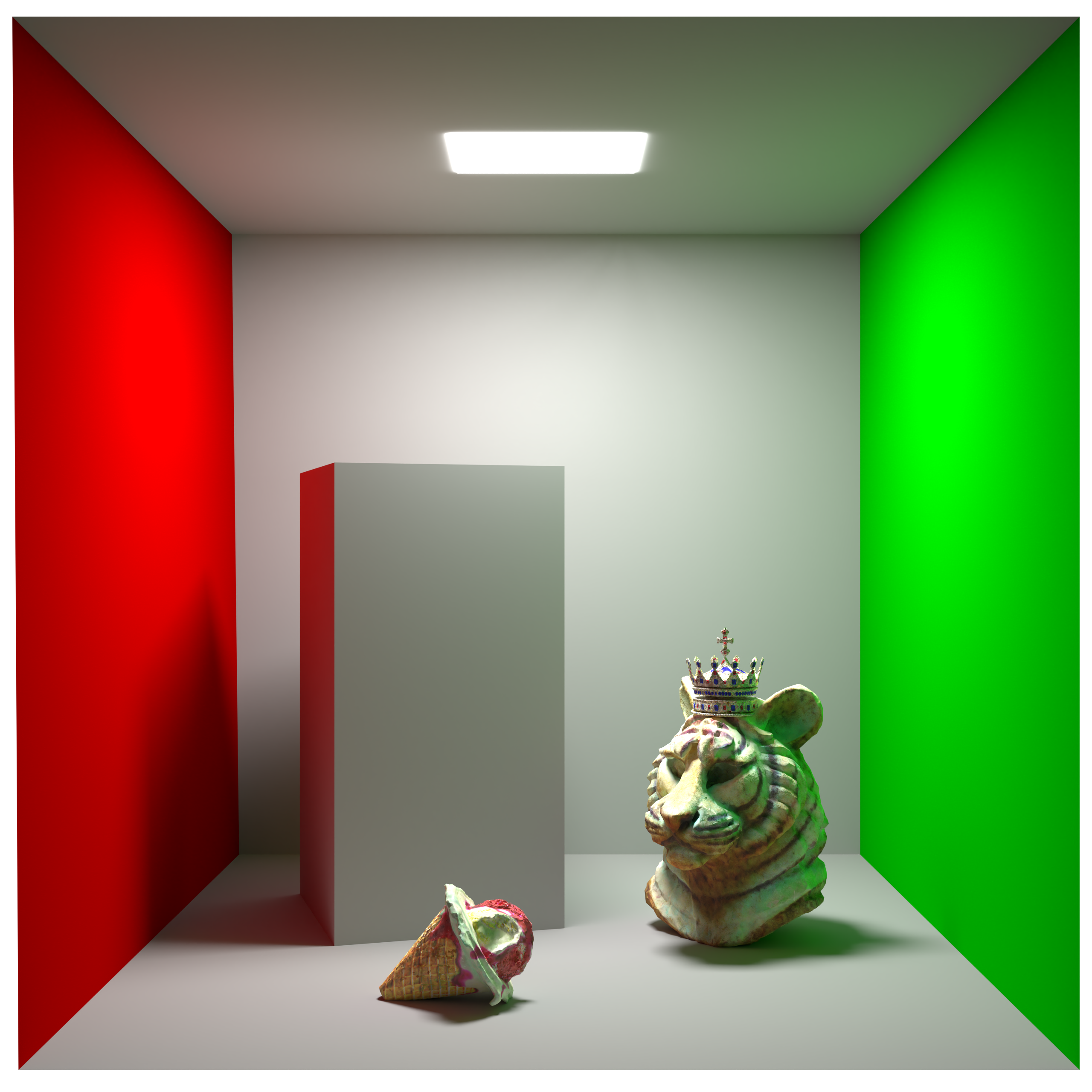}
\end{center}
\vspace{-1em}
   \caption{\textbf{Scene Editing.} Three generated objects, namely crown, tiger, and ice cream, are imported into the Cornell Box scene. The scene is then rendered using the Cycles path tracer in Blender, producing natural shadows and reflectance effects.}
\label{fig: scene editing}
\end{figure}

\begin{figure}
\includegraphics[width=0.48\textwidth]{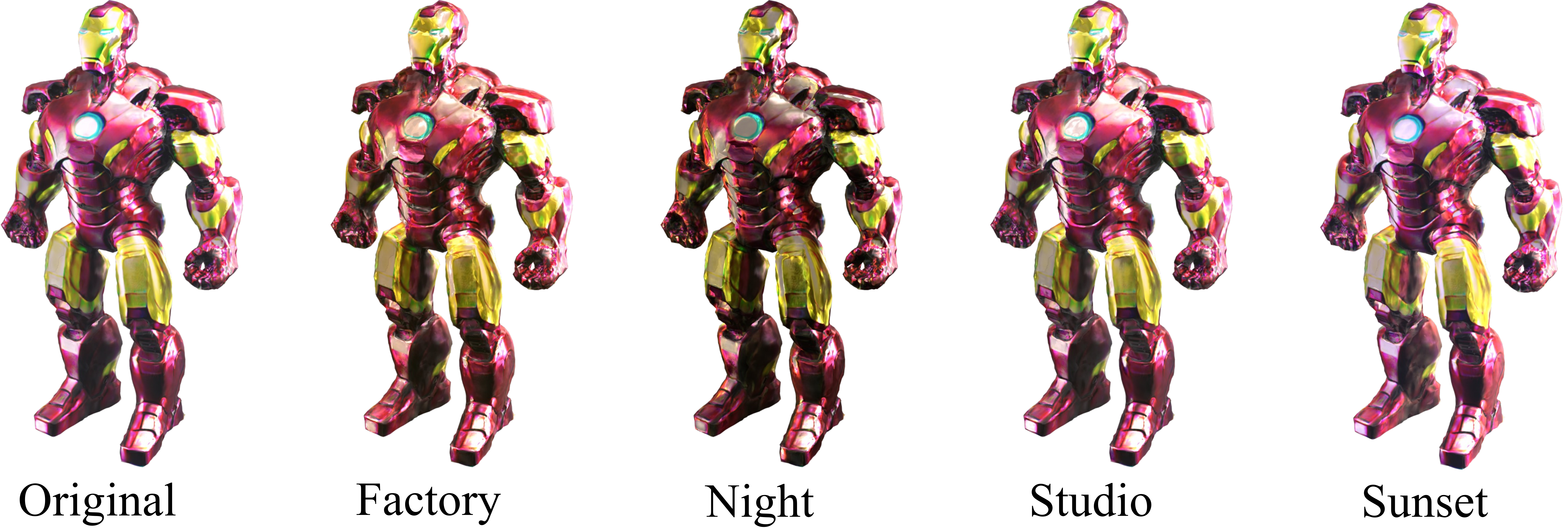}
   \caption{\textbf{Relighting.} Our generated Iron Man model is relit using different lighting setups in Blender, producing diverse reflectance effects on the armor.}
\label{fig:relighting_results}
\end{figure}

\subsection{Scene editing and simulation}
\label{exp:scene_editing}

Given the disentangled design, our method produces high-quality generation of both the surface and appearance, and is compatible with popular graphics engines for scene editing and physical simulation. In Figure \ref{fig:head}, we import into Blender a stone statue generated by our method from the text prompt of "a highly detailed stone bust of Theodoros Kolokotronis", where soft body and cloth simulations are performed along with material editing. Given the high quality of the generated surface geometry (e.g., no holes on the surface), the simulations are of high-level physical accuracy, as shown in the accompanying supplemental video. Moreover, we showcase the ability to modify the material of the statue by replacing it with another wood material downloaded from the Internet \cite{poliigon}, which poses a challenge for comparative methods such as DreamFusion \cite{poole2022dreamfusion}. Additionally, in Fig. \ref{fig: scene editing}, by importing generated ice cream, tiger, and crown into the Cornell Box, we demonstrate the plausible physical interaction between our generated results and the scene, with natural shadows being cast. Finally, Fig. \ref{fig:relighting_results} illustrates the replacement of the HDR environment map to produce diverse lighting and corresponding reflectance effects on the generated iron man.

\section{Limitations}
While Fantasia3D demonstrates promising advancements in the realm of generating photorealistic 3D assets from text prompts, several limitations remain.
For instance, while our method successfully produces loose visual effects, it remains a significant challenge to generate corresponding loose geometries, such as hair, fur, and grass, solely based on text prompts.  Additionally, our method primarily emphasizes object generation, thereby lacking the capacity to generate complete scenes with background from text prompts. Consequently, our future research endeavors will be dedicated to addressing these limitations by focusing on the generation of complete scenes and intricate loose geometries.


\section{Conclusion}

In this paper, we present Fantasia3D, a new method for automatic text-to-3D generation. Fantasia3D uses disentangled modeling and learning of geometry and appearance, and is able to generate both the fine surface and rich material/texture. Fantasia3D is based on the hybrid scene representation of \textsc{DMTet}. For geometry learning, we propose to encode a rendered normal map, and use shape encoding of the normal as the input of the pre-trained, stable diffusion model. For appearance modeling, we introduce the spatially varying BRDF into the text-to-3D task, thus enabling material learning that supports photorealistic rendering of the learned surface. Expect for text prompts, our method can be triggered with a customized 3D shape as well; this is flexible for users to better control what content is to be generated. Our method is also convenient to support relighting, editing, and physical simulation of the generated 3D assets. Our method is based on pre-trained image diffusion models (i.e., the stable diffusion). In future research, we are interested in learning 3D diffusion directly from the large language models. 

\justify \noindent \textbf{Acknowledgements.} This work is supported in part by Program for Guangdong Introducing Innovative and Entrepreneurial Teams (No.: 2017ZT07X183) and Guangdong R\&D key project of China (No.: 2019B010155001).
\newpage
{\small
\bibliographystyle{ieee_fullname}
\bibliography{egbib}
}

\end{document}